\documentclass[conference]{IEEEtran}
\IEEEoverridecommandlockouts
\usepackage{cite}
\usepackage{amsmath,amssymb,amsfonts}
\usepackage{algorithm}
\usepackage{algorithmic}
\usepackage{graphicx}
\usepackage{booktabs}
\usepackage[table]{xcolor}
\usepackage{tikz}
\usetikzlibrary{positioning,arrows.meta,fit,calc}
\usepackage[hidelinks]{hyperref}

\definecolor{maroon}{HTML}{7A1F2B}
\definecolor{rose}{HTML}{F2ECED}
\definecolor{gain}{HTML}{F5DADE}

\makeatletter
\newcommand{\linebreakand}{%
  \end{@IEEEauthorhalign}
  \hfill\mbox{}\par
  \mbox{}\hfill\begin{@IEEEauthorhalign}
}
\makeatother

\begin{document}

\title{\fontsize{20}{24}\selectfont Certified Defense against Adversarial Patches\\ on Traffic Signs via Mask-Aware Training}

\author{\IEEEauthorblockN{Rudra Saxena}
\IEEEauthorblockA{\textit{Vellore Institute of Technology}\\
Chennai, India \\
rudra.saxena2023@vitstudent.ac.in}
\and
\IEEEauthorblockN{Vishal Dev Verma}
\IEEEauthorblockA{\textit{Vellore Institute of Technology}\\
Chennai, India \\
vishaldev.verma2023@vitstudent.ac.in}
\and
\IEEEauthorblockN{Rohan Sharma}
\IEEEauthorblockA{\textit{Vellore Institute of Technology}\\
Chennai, India \\
rohan.sharma2023@vitstudent.ac.in}
\linebreakand
\IEEEauthorblockN{Dr.\ Palani Thanaraj K}
\IEEEauthorblockA{\textit{Vellore Institute of Technology}\\
Chennai, India \\
palanithanaraj.k@vit.ac.in}}

\maketitle

\begin{abstract}
Adversarial patches are a practical threat to traffic-sign recognition: a small sticker on a sign can
change a classifier's decision. PatchCleanser is a certified defense that applies two rounds of masking
and, for every image it certifies, provably returns the correct label for any patch that fits inside one mask. We show that on
traffic signs the defense certifies almost nothing when paired with a normally trained classifier
(1.7\% to 4.8\% certified accuracy on GTSRB), because a single mask can hide the small central symbol that
distinguishes sign classes. We propose \emph{mask-aware training}, which trains the classifier on
images occluded by random pairs drawn from the defense's own mask set, with a sampling bias toward
masks over the sign's centre. The defense, mask set and certification procedure are unchanged, so the
guarantee is preserved. For an ImageNet-pretrained ResNet-50, mask-aware training raises certified
accuracy from 4.8\% to 60.0\%. Under real per-image adversarial patches, the defended ResNet-50 reaches
85.7\% accuracy, versus 74.6\% for the original PatchCleanser with a normally trained ResNet-50.
\end{abstract}


\section{Introduction}
Deep classifiers used in driver assistance and autonomous driving read traffic signs from camera
images. Physical adversarial patches, small printed stickers placed on or near an object, can cause
confident misclassification~\cite{brown2017patch,eykholt2018robust}. Empirical defenses are routinely
broken by adaptive attacks, which motivates \emph{certified} defenses: methods that prove a
prediction cannot be changed by any patch of a given size~\cite{chiang2020certified,levine2020randomized,xiang2021patchguard}.

PatchCleanser~\cite{xiang2022patchcleanser} is a certified defense that works with any image
classifier. It evaluates the classifier on images occluded by one and then two masks from an
\emph{R-covering} mask set and certifies an image when all two-mask predictions agree with the label.

Traffic signs are harder. A sign's class is decided by a small symbol at its centre (the digits of
a speed limit, the pictogram of a warning sign), and a mask sized to cover a patch can hide that symbol
entirely. We find that PatchCleanser with a normally trained model certifies only 1.7\% to 4.8\% of
GTSRB test images. This paper makes the following contributions:
\begin{itemize}
\item An end-to-end certified-patch-defense pipeline for traffic signs (GTSRB), including training,
per-image adversarial patch generation, two-mask inference and certification.
\item \emph{Mask-aware training}: training with random pairs from the defense's own mask set, biased
toward the grid centre. It requires no change to inference and keeps the certificate valid.
\item A comparison with the original PatchCleanser on a wide CNN and a ResNet-50: certified accuracy
rises from 1.7--4.8\% to 53.5--60.0\%, and accuracy under real adversarial patches from 72.8--74.6\%
to 80.9--85.7\%.
\end{itemize}

\begin{figure*}[t]
\centering
\includegraphics[width=\textwidth]{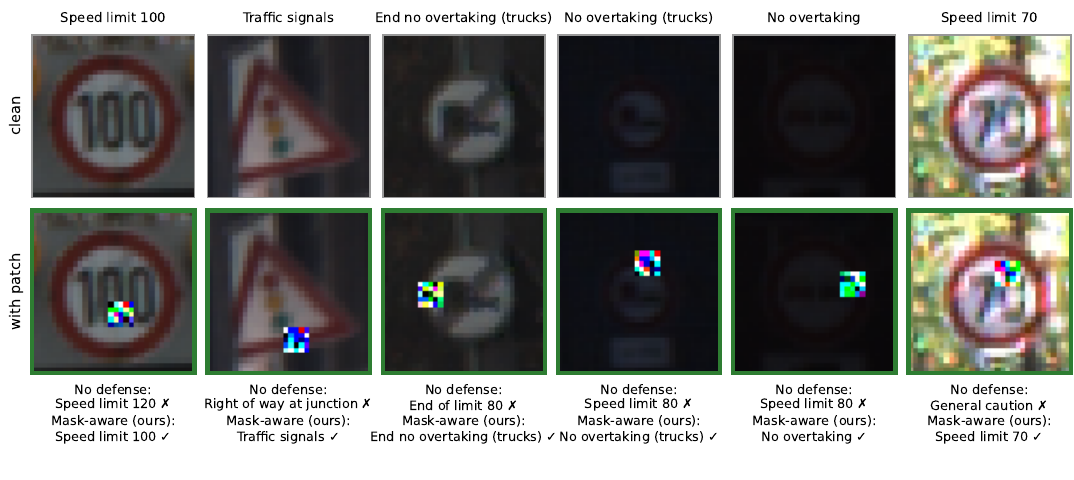}
\caption{Examples of GTSRB test signs (top) and the same signs with an adversarial $5\times5$ patch
(bottom), using the wide CNN trained with our method (Table~\ref{tab:setup}). Without a defense the classifier
is fooled (e.g., Speed limit 100 $\to$ 120); with mask-aware training (ours) inside the unchanged
double-masking defense it returns the correct class.}
\label{fig:demo}
\end{figure*}

\section{Background and Related Work}
\subsubsection{Adversarial patches}
Brown et al.~\cite{brown2017patch} introduced universal, printable adversarial patches, and Eykholt
et al.~\cite{eykholt2018robust} demonstrated physical sticker attacks that cause stop signs to be misread.
The attacker controls all pixels inside a contiguous region and nothing outside it.

\subsubsection{Certified patch defenses}
Interval bound propagation~\cite{chiang2020certified}, derandomized smoothing~\cite{levine2020randomized}
and small-receptive-field models~\cite{xiang2021patchguard} give provable guarantees but constrain the
architecture. PatchCleanser~\cite{xiang2022patchcleanser} is architecture-agnostic, and later work
improves its efficiency and utility~\cite{xiang2024patchcure,li2022vip}. Saha et al.~\cite{saha2023revisiting}
show that occlusion-based training (Greedy Cutout) improves PatchCleanser on natural-image benchmarks;
PatchCleanser itself fine-tunes with Cutout~\cite{devries2017cutout}. We study the traffic-sign
domain, where the defense fails without such training, and train with the defense's own mask pairs.

\subsubsection{Double masking}
Let $f$ be a classifier and $x$ an image. A mask $m\in\{0,1\}^{H\times W}$ zeroes a square region.
A mask set $\mathcal{M}=\{m_1,\dots,m_k\}$ is \emph{R-covering} for a $p\times p$ patch if every possible
patch location is fully covered by at least one mask. With stride $s$ and mask size $p+s-1$, a
$\sqrt{k}\times\sqrt{k}$ grid of masks is R-covering~\cite{xiang2022patchcleanser}. The defense
computes one-mask predictions $\hat y_i=f(x\odot m_i)$ and, when they disagree, two-mask predictions
$\hat y_{ij}=f(x\odot m_i\odot m_j)$ (Algorithm~\ref{alg:dm}). If $\hat y_{ij}=y$ for all $i,j$ on the
clean image, the output equals $y$ for every patched version of $x$~\cite{xiang2022patchcleanser}; this is
the certificate (Algorithm~\ref{alg:cert}).

\section{Method}

\subsection{Threat Model and Mask Set}
\begin{figure}[t]
\centering
\resizebox{\columnwidth}{!}{%
\begin{tikzpicture}[
  font=\small, node distance=4mm and 5mm,
  box/.style={draw=maroon, fill=rose, rounded corners=2pt, align=center, minimum height=10mm, inner sep=2.5pt, text width=#1},
  box/.default=20mm,
  arr/.style={-{Stealth[length=2mm]}, draw=maroon, line width=0.7pt}]
\node[box=17mm] (tr) {Training\\ images};
\node[box=23mm, right=of tr] (samp) {Sample mask pair $(i,j)$, centre-biased};
\node[box=21mm, right=of samp] (occ) {Occlude\\ $x\odot m_i\odot m_j$};
\node[box=17mm, right=of occ, fill=maroon, text=white] (f) {Classifier\\ $f_\theta$};
\draw[arr] (tr) -- (samp); \draw[arr] (samp) -- (occ); \draw[arr] (occ) -- (f);
\node[box=17mm, below=10mm of tr] (te) {Test image\\ (+ patch)};
\node[box=23mm, right=of te] (one) {36 one-mask predictions $\hat y_i$};
\node[box=21mm, right=of one] (two) {Up to 666 two-mask predictions $\hat y_{ij}$};
\node[box=17mm, right=of two] (out) {Label +\\ certificate};
\draw[arr] (te) -- (one); \draw[arr] (one) -- (two); \draw[arr] (two) -- (out);
\draw[arr, dashed] (f.south) -- ++(0,-5mm) -| node[pos=0.25, above, font=\footnotesize, text=maroon] {trained $f_\theta$} ($(one.north)!0.5!(two.north)$);
\node[font=\footnotesize\bfseries, text=maroon, anchor=south west] at ($(tr.north west)+(0,0.8mm)$) {(a) Mask-aware training};
\node[font=\footnotesize\bfseries, text=maroon, anchor=north west] at ($(te.south west)+(0,-0.8mm)$) {(b) PatchCleanser inference and certification (unchanged)};
\end{tikzpicture}}
\caption{Pipeline. (a) The classifier is trained on images occluded by pairs from the defense's own
R-covering mask set. (b) Inference and certification use the unmodified double-masking defense.}
\label{fig:overview}
\end{figure}
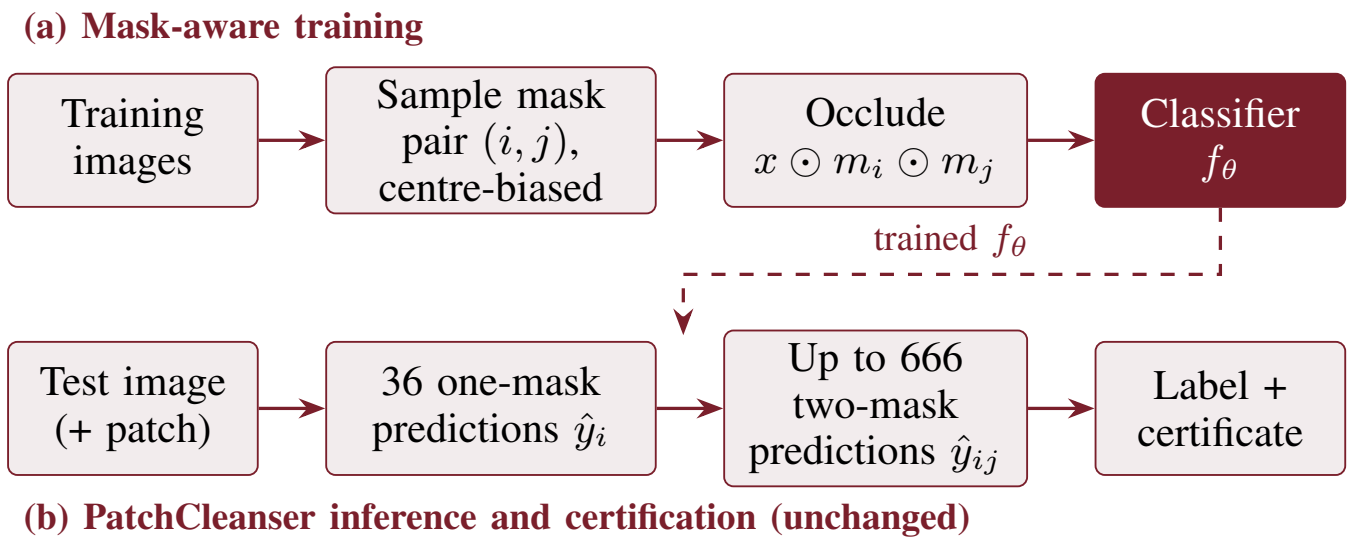

The attacker places one square patch covering 2.4\% of the image pixels at any location with
arbitrary content: $5\times5$ on $32\times32$ inputs and $20\times20$ on $128\times128$ inputs. We use a
$6\times6$ grid ($k=36$) of $9\times9$ masks at stride 5 (respectively $38\times38$ at stride 19), which is
R-covering for the patch. A two-mask prediction uses an unordered pair $\{i,j\}$ of masks, with $i=j$
meaning a single mask, so there are $\binom{36}{2}+36=630+36=666$ of them, i.e.\ $k(k+1)/2$.
Certification requires all 666 to be correct;
inference needs at most 666, since Algorithm~\ref{alg:dm} evaluates two-mask predictions only for
disagreeing masks. We compute all 666 once per image and reuse them for both.
Fig.~\ref{fig:overview} summarizes the pipeline.

\begin{figure}[t]
\centering
\includegraphics[width=\columnwidth]{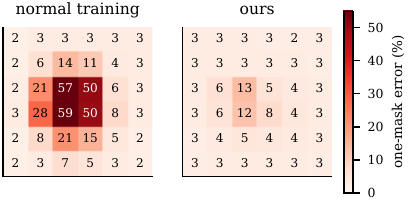}
\caption{Error (\%) when a single mask is applied, by mask position on the $6\times6$ grid. Left: the
wide CNN with normal training; right: the same network with mask-aware training (ours), which removes
the central failure mode.}
\label{fig:heatmap}
\end{figure}

\subsection{Why Traffic Signs Are Hard to Certify}
Fig.~\ref{fig:heatmap} shows the error of the normally trained wide CNN when one mask is applied. Masks at
the border cost about 3\%, but the four masks over the centre raise the error to 50--59\%, and none of
the 651 speed-limit test images is certified: once the digits are hidden, 30, 50 and 80 look the same. Because
certification needs all 666 two-mask predictions to be correct, these central failures are enough to
break the certificate on almost every image.

\subsection{Mask-Aware Training}
We train the classifier to predict the correct label from images occluded by pairs of the defense's
own masks (Algorithm~\ref{alg:train}):
\begin{equation}
\min_\theta\; \mathbb{E}_{(x,y)}\,\mathbb{E}_{(i,j)\sim\pi}\;\ell\big(f_\theta(x\odot m_i\odot m_j),\,y\big),
\end{equation}
where $\ell$ is cross-entropy. The sampling distribution $\pi$ draws $j$ uniformly and, with probability
$\tfrac12$, draws $i$ from the interior masks $\mathcal{I}$ of the grid (the $4\times4$ centre), otherwise
uniformly. A fraction $q$ of each batch is left unmasked. In our models every training image
is masked ($q=0$). Since inference and
certification are not modified, the PatchCleanser guarantee holds for the trained model.

\setlength{\floatsep}{5pt}
\setlength{\textfloatsep}{7pt}
\begin{algorithm}[!t]
\caption{Mask-aware training}
\label{alg:train}
\begin{algorithmic}[1]
\REQUIRE data $\mathcal{D}$, mask set $\mathcal{M}$, interior set $\mathcal{I}\subset\mathcal{M}$, keep rate $q$, epochs $E$
\ENSURE classifier $f_\theta$
\FOR{$e=1$ \TO $E$}
  \FOR{each batch $\{(x_b,y_b)\}\subset\mathcal{D}$}
    \STATE $x_b \leftarrow x_b\cdot u_b$, \ $u_b\sim\mathcal{U}(0.7,1.3)$ \COMMENT{brightness}
    \STATE $j_b \sim \mathcal{U}(\mathcal{M})$
    \STATE $i_b \sim \mathcal{U}(\mathcal{I})$ w.p.\ $\tfrac12$, else $i_b\sim\mathcal{U}(\mathcal{M})$
    \STATE $\tilde x_b \leftarrow x_b$ w.p.\ $q$, else $x_b\odot m_{i_b}\odot m_{j_b}$
    \STATE $\theta \leftarrow \theta - \eta\nabla_\theta \frac{1}{B}\sum_b \ell(f_\theta(\tilde x_b),y_b)$
  \ENDFOR
\ENDFOR
\end{algorithmic}
\end{algorithm}

\begin{algorithm}[!t]
\caption{Double-masking inference~\cite{xiang2022patchcleanser}}
\label{alg:dm}
\begin{algorithmic}[1]
\REQUIRE image $x$, classifier $f$, mask set $\mathcal{M}$
\ENSURE label $\hat y$
\STATE $\hat y_i \leftarrow f(x\odot m_i)$ for $i=1..k$
\IF{all $\hat y_i$ are equal}
  \RETURN $\hat y_1$ \COMMENT{Case I: agreement}
\ENDIF
\STATE $y_{\text{maj}} \leftarrow$ majority of $\{\hat y_i\}$
\FOR{each $i$ with $\hat y_i\neq y_{\text{maj}}$}
  \IF{$f(x\odot m_i\odot m_j)=\hat y_i$ for all $j$}
    \RETURN $\hat y_i$ \COMMENT{Case II: consistent disagreer}
  \ENDIF
\ENDFOR
\RETURN $y_{\text{maj}}$ \COMMENT{Case III}
\end{algorithmic}
\end{algorithm}

\begin{algorithm}[!t]
\caption{Certification~\cite{xiang2022patchcleanser}}
\label{alg:cert}
\begin{algorithmic}[1]
\REQUIRE clean image $x$, label $y$, classifier $f$, mask set $\mathcal{M}$
\FOR{$1\le i\le j\le k$}
  \IF{$f(x\odot m_i\odot m_j)\neq y$}
    \RETURN not certified
  \ENDIF
\ENDFOR
\RETURN certified \COMMENT{no patch can change the output}
\end{algorithmic}
\end{algorithm}

\section{Experimental Setup}
\textbf{Dataset.} GTSRB~\cite{stallkamp2012man}: 26{,}640 training and 12{,}630 test images, 43 classes.
We evaluate on a fixed random subset of 2{,}000 test images; every method sees the same images and
patch locations.

\textbf{Models.} Table~\ref{tab:setup} lists the classifiers. The wide CNN (five convolutional layers)
is trained from scratch at $32\times32$. ResNet-50~\cite{he2016deep} starts from ImageNet weights and is
fine-tuned at $128\times128$ on a Colab T4 GPU with mixed precision. For each model, the original
PatchCleanser uses the normally trained network and ours uses the mask-aware one; the defense is identical.

\textbf{Attack.} \emph{Adversarial patch}: per-image, untargeted, 100 Adam~\cite{kingma2015adam} steps on the patch pixels maximizing the
cross-entropy of the undefended model at a random location. Certified accuracy covers every patch
content and location in the threat model.

\section{Results}
\begin{table}[!t]
\caption{Model Architectures and Parameters}
\label{tab:setup}
\centering\footnotesize
\setlength{\tabcolsep}{4pt}
\begin{tabular}{@{}lrrrc@{}}
\toprule
Model & Params & Input & Patch & Epochs\\[-1pt]
& & (px) & (px) & \\
\midrule
Wide CNN (from scratch)  & 0.73M & 32  & 5  & 40\\
ResNet-50 (pretrained)   & 23.6M & 128 & 20 & 30\\
\bottomrule
\end{tabular}

\vspace{2pt}
{\scriptsize Each model is trained twice with the same recipe: normally (original PatchCleanser) and with mask-aware training (ours). Both wide CNNs also use a random $\pm2$-pixel translation during training.\par}
\end{table}

\begin{table}[!t]
\caption{Certified and Defended Accuracy}
\label{tab:main}
\centering\footnotesize
\setlength{\tabcolsep}{3.5pt}
\begin{tabular}{@{}lccccc@{}}
\toprule
& \multicolumn{3}{c}{Clean images} & \multicolumn{2}{c}{Adversarial patch}\\
\cmidrule(lr){2-4}\cmidrule(lr){5-6}
Method & No def. & Defense & \textbf{Certified} & No def. & Defense\\
\midrule
\multicolumn{6}{@{}l}{\textbf{Wide CNN}}\\
\hspace{4pt}PatchCleanser & 97.9 & 93.8 & 1.7 & 76.1 & 72.8\\
\hspace{4pt}Ours          & 97.3 & 93.2 & \cellcolor{gain}\textbf{53.5} & 70.0 & \cellcolor{gain}\textbf{80.9}\\
\addlinespace[1.5pt]
\multicolumn{6}{@{}l}{\textbf{ResNet-50}}\\
\hspace{4pt}PatchCleanser & 99.2 & 96.1 & 4.8 & 76.6 & 74.6\\
\hspace{4pt}Ours          & 97.9 & 93.7 & \cellcolor{gain}\textbf{60.0} & 80.7 & \cellcolor{gain}\textbf{85.7}\\
\bottomrule
\end{tabular}

\vspace{2pt}
{\scriptsize Original PatchCleanser vs.\ ours on 2{,}000 GTSRB test images (\%). Shaded cells highlight our gains.\par}
\end{table}

\subsection{Certified Accuracy}
\begin{figure}[!b]
\centering
\includegraphics[width=\columnwidth]{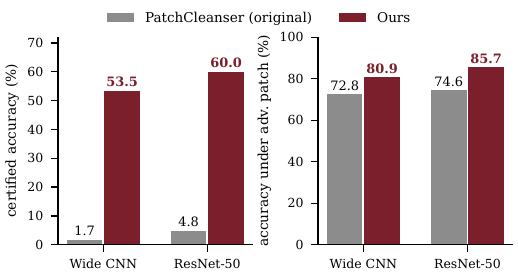}
\caption{Original PatchCleanser vs.\ ours on GTSRB (2.4\% patch). Left: certified accuracy; right:
accuracy with the defense under a real adversarial patch.}
\label{fig:cert}
\end{figure}

Table~\ref{tab:main} and Fig.~\ref{fig:cert} compare the original PatchCleanser with ours. The original
certifies only 1.7\% (wide CNN) and 4.8\% (ResNet-50) of test images. With mask-aware training the same
defense certifies 53.5\% and 60.0\%, gains of 51.8 and 55.2 points. Clean accuracy with the defense
drops by 0.6 and 2.4 points.

\subsection{Robustness to Real Adversarial Patches}
Without a defense, adversarial patches reduce accuracy to 70.0--80.7\% (Table~\ref{tab:main}). The
original PatchCleanser makes this slightly worse (72.8\% vs.\ 76.1\% and 74.6\% vs.\ 76.6\%), because
the normally trained model misclassifies the masked images. With ours the defended accuracy reaches
80.9\% and 85.7\%, 8.1 and 11.1 points above the original. Fig.~\ref{fig:demo} shows examples.

\subsection{Limitations}
We use 2{,}000 of the 12{,}630 test images (666 forward passes per image). The adversarial patch
targets the undefended model, not the defense; certified accuracy holds against any patch. Each
setting is trained once, and all patches are digital.

\section{Conclusion}
PatchCleanser's guarantee transfers to traffic-sign recognition only if the classifier can recognize
signs with a mask over their central symbol. Mask-aware training with the defense's own mask pairs and
a centre bias supplies this ability without modifying the defense. On GTSRB our best model, a
ResNet-50, goes from 4.8\% to 60.0\% certified and reaches 85.7\% defended accuracy under adversarial
patches. Future work
includes the full test set and physical stickers.

\section*{Appendix A --- Terminology and Definitions}
\footnotesize
\setlength{\parskip}{0.5pt}
\noindent\textbf{GTSRB}: The German Traffic Sign Recognition Benchmark~\cite{stallkamp2012man} contains 39{,}270 natural images of 43 road sign classes (26{,}640 train, 12{,}630 test) across diverse scales, lighting, and degradation. We evaluate on a fixed subset of 2{,}000 test images with identical patch coordinates across all models.\par
\noindent\textbf{Adversarial patch}: A localized perturbation~\cite{brown2017patch,eykholt2018robust} where an adversary freely modifies pixels within a bounded spatial region (e.g., a physical sticker on a sign), leaving other pixels intact. In our threat model, the patch covers 2.4\% of pixels ($5\times5$ on $32\times32$, $20\times20$ on $128\times128$).\par
\noindent\textbf{Threat model}: Formal adversary definition: placing one square patch of up to 2.4\% pixel area anywhere on the sign with arbitrary continuous RGB values to cause misclassification, subject to no $L_p$-norm bound.\par
\noindent\textbf{Defense}: An algorithmic pipeline wrapped around a classifier to resist attacks. Here, we evaluate PatchCleanser, a certified defense based on two rounds of spatial masking.\par
\noindent\textbf{PatchCleanser}: An architecture-agnostic certified defense~\cite{xiang2022patchcleanser} that guarantees robustness against localized patches via double masking without modifying model weights.\par
\noindent\textbf{Mask}: A spatial occlusion operator $m\in\{0,1\}^{H\times W}$ setting pixels to zero in a square window ($x\odot m$) prior to inference ($9\times9$ for $32\times32$; $38\times38$ for $128\times128$), suppressing adversarial patches.\par
\noindent\textbf{Mask set, $k$}: A discrete set $\mathcal{M}=\{m_1,\dots,m_k\}$ of $k=36$ masks arranged on a uniform $6\times6$ grid across the image, governing spatial resolution and evaluation overhead.\par
\noindent\textbf{R-covering}: The structural property ensuring every valid $p\times p$ patch location is fully enclosed by at least one mask. Sizing masks to $p+s-1$ for stride $s$ guarantees covering across continuous offsets.\par
\noindent\textbf{One-/two-mask prediction}: Model inferences on occluded inputs: $\hat y_i = f(x\odot m_i)$ for single mask $m_i$; $\hat y_{ij} = f(x\odot m_i\odot m_j)$ for mask pair $\{m_i,m_j\}$ ($i=j$ denotes single mask).\par
\noindent\textbf{666 (Two-mask evaluations)}: Total count of unordered mask pairs from $k=36$ masks, including pairing a mask with itself: $\binom{36}{2}+36=\frac{36\times37}{2}=666$, evaluated once per sample and shared.\par
\noindent\textbf{Double masking}: PatchCleanser's inference rule (Algorithm~\ref{alg:dm}). It checks consensus among one-mask predictions (Case I) and identifies if a minority prediction consistently disagrees across secondary masks (Case II); otherwise majority vote is returned (Case III).\par
\noindent\textbf{Certificate}: Formal verification that no patch in the threat model can change the defended output on a clean image (Algorithm~\ref{alg:cert}), proven when all 666 two-mask evaluations match label $y$.\par
\noindent\textbf{Certified accuracy}: Fraction of test images correctly classified and certified, establishing an attack-independent lower bound on defended accuracy against any admissible patch.\par
\noindent\textbf{No defense / Defense}: Comparative evaluation settings: \emph{No defense} measures raw classifier accuracy; \emph{Defense} measures accuracy when wrapped inside the double-masking pipeline.\par
\noindent\textbf{Mask-aware training}: Our training method (Algorithm~\ref{alg:train}) optimizing the classifier on images occluded by sampled pairs from $\mathcal{M}$, conditioning it to recognize signs under dual occlusions.\par
\noindent\textbf{Centre bias}: A sampling distribution $\pi$ during training where mask $i$ is drawn from interior masks $\mathcal{I}\subset\mathcal{M}$ ($4\times4$) with probability 0.5 (else uniformly), forcing reliance on peripheral cues.\par
\noindent\textbf{$q$ (keep rate)}: The fraction of each training minibatch left unmasked. We set $q=0$, occluding every training image to maximize robustness to severe occlusions.\par
\noindent\textbf{Normal training}: Standard training on clean images with brightness scaling only; serves as the baseline model in original PatchCleanser.\par
\noindent\textbf{PatchCleanser (original)}: Baseline combining the unmodified PatchCleanser defense with a normally trained neural network.\par
\noindent\textbf{Ours}: The combination of the exact PatchCleanser defense and certification procedure with a classifier trained via mask-aware training.\par
\noindent\textbf{Wide CNN}: A five-layer convolutional neural network (0.73M parameters) trained from scratch on $32\times32$ inputs with $\pm2$-pixel random translations and brightness augmentations for 40 epochs.\par
\noindent\textbf{ResNet-50}: A 50-layer residual network (23.6M parameters)~\cite{he2016deep} initialized with ImageNet weights and fine-tuned on GTSRB at $128\times128$ for 30 epochs using mixed precision on GPU.\par
\setlength{\parskip}{0pt}

\end{document}